\documentclass[twoside,twocolumn,9pt]{article}
\usepackage{extsizes}
\usepackage[super,sort&compress,comma]{natbib} 
\usepackage[version=3]{mhchem}
\usepackage[left=1.5cm, right=1.5cm, top=1.785cm, bottom=2.0cm]{geometry}
\usepackage{balance}
\usepackage{mathptmx}
\usepackage{sectsty}
\usepackage{graphicx} 
\usepackage{lastpage}
\usepackage[format=plain,justification=justified,singlelinecheck=false,font={stretch=1.125,small,sf},labelfont=bf,labelsep=space]{caption}
\usepackage{float}
\usepackage{fancyhdr}
\usepackage{fnpos}
\usepackage[english]{babel}
\addto{\captionsenglish}{%
  
}
\usepackage{array}
\usepackage{droidsans}
\usepackage{charter}
\usepackage[T1]{fontenc}
\usepackage[usenames,dvipsnames]{xcolor}
\usepackage{setspace}
\usepackage[compact]{titlesec}
\usepackage{hyperref}

\usepackage{minted}
\usepackage{tcolorbox}
\tcbuselibrary{listings,breakable}
\usepackage{fancyvrb}
\usepackage{amsmath,amssymb}
\usepackage[x11names]{xcolor} %
\usepackage{wrapfig}
\usepackage{hyperref}
\usepackage{url}

\floatstyle{ruled}
\newfloat{code}{thp}{lop}
\floatname{code}{Code Block}

\usepackage{epstopdf}

\definecolor{cream}{RGB}{222,217,201}

\begin{document}

\pagestyle{fancy}
\thispagestyle{plain}
\fancypagestyle{plain}{
\renewcommand{\headrulewidth}{0pt}
}

\makeFNbottom
\makeatletter
\renewcommand\LARGE{\@setfontsize\LARGE{15pt}{17}}
\renewcommand\Large{\@setfontsize\Large{12pt}{14}}
\renewcommand\large{\@setfontsize\large{10pt}{12}}
\renewcommand\footnotesize{\@setfontsize\footnotesize{7pt}{10}}
\makeatother

\renewcommand{\thefootnote}{\fnsymbol{footnote}}
\renewcommand\footnoterule{\vspace*{1pt}%
\color{cream}\hrule width 3.5in height 0.4pt \color{black}\vspace*{5pt}} 
\setcounter{secnumdepth}{5}

\makeatletter 
\renewcommand\@biblabel[1]{#1}            
\renewcommand\@makefntext[1]%
{\noindent\makebox[0pt][r]{\@thefnmark\,}#1}
\makeatother 
\renewcommand{\figurename}{\small{Fig.}~}
\sectionfont{\sffamily\Large}
\subsectionfont{\normalsize}
\subsubsectionfont{\bf}
\setstretch{1.125} 
\setlength{\skip\footins}{0.8cm}
\setlength{\footnotesep}{0.25cm}
\setlength{\jot}{10pt}
\titlespacing*{\section}{0pt}{4pt}{4pt}
\titlespacing*{\subsection}{0pt}{15pt}{1pt}

\fancyfoot{}
\fancyfoot[LO,RE]{\vspace{-7.1pt}\includegraphics[height=9pt]{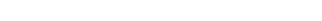}}
\fancyfoot[CO]{\vspace{-7.1pt}\hspace{13.2cm}\includegraphics{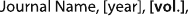}}
\fancyfoot[CE]{\vspace{-7.2pt}\hspace{-14.2cm}\includegraphics{head_foot/RF}}
\fancyfoot[RO]{\footnotesize{\sffamily{1--\pageref{LastPage} ~\textbar  \hspace{2pt}\thepage}}}
\fancyfoot[LE]{\footnotesize{\sffamily{\thepage~\textbar\hspace{3.45cm} 1--\pageref{LastPage}}}}
\fancyhead{}
\renewcommand{\headrulewidth}{0pt} 
\renewcommand{\footrulewidth}{0pt}
\setlength{\arrayrulewidth}{1pt}
\setlength{\columnsep}{6.5mm}
\setlength\bibsep{1pt}

\makeatletter 
\newlength{\figrulesep} 
\setlength{\figrulesep}{0.5\textfloatsep} 

\newcommand{\topfigrule}{\vspace*{-1pt}%
\noindent{\color{cream}\rule[-\figrulesep]{\columnwidth}{1.5pt}} }

\newcommand{\botfigrule}{\vspace*{-2pt}%
\noindent{\color{cream}\rule[\figrulesep]{\columnwidth}{1.5pt}} }

\newcommand{\dblfigrule}{\vspace*{-1pt}%
\noindent{\color{cream}\rule[-\figrulesep]{\textwidth}{1.5pt}} }

\makeatother

\twocolumn[
  \begin{@twocolumnfalse}
{\includegraphics[height=30pt]{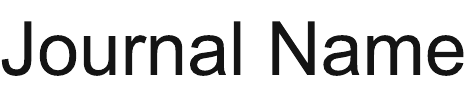}\hfill\raisebox{0pt}[0pt][0pt]{\includegraphics[height=55pt]{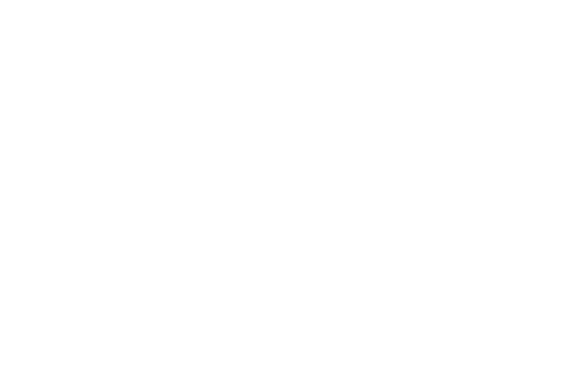}}\\[1ex]
\includegraphics[width=18.5cm]{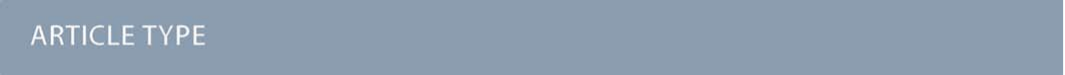}}\par
\vspace{1em}
\sffamily
\begin{tabular}{m{4.5cm} p{13.5cm} }

\includegraphics{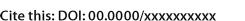} & \noindent\LARGE{\textbf{An Agentic Orchestration of Atomistic Simulations}} \\
\vspace{0.3cm} & \vspace{0.3cm} \\

 & \noindent\large{Rahul Somasundaram,
 \textit{$^{a\dag}$} Adela Habib,\textit{$^{a}$} Khanh Dang,\textit{$^{a}$} Sachin Shivakumar,\textit{$^{a}$} Ryley G. Hill,\textit{$^{a}$} Golo Wimmer,\textit{$^{a}$} Avanish Mishra,\textit{$^{a}$} Aleksandra Pachalieva,\textit{$^{a}$} Arthur Lui,\textit{$^{a}$} Hari Viswanathan,\textit{$^{a}$} Michael Grosskopf,\textit{$^{a}$} Saryu Fensin,\textit{$^{a}$} Russell Bent,\textit{$^{a}$} Nathan DeBardeleben,\textit{$^{a}$} and Earl Lawrence\textit{$^{a}$}} \\

\includegraphics{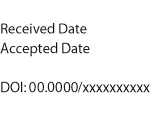} & \noindent\normalsize{Atomistic simulations are central to materials design, but their execution involves complex, multi-step workflows that require significant human expertise. Here, we present an agent-based system embedded within the URSA (Universal Research and Scientific Agent) framework that automates the design, execution, and validation of atomistic simulations, demonstrated using the  Large-scale Atomic/Molecular Massively Parallel Simulator (LAMMPS) tool. Our system autonomously selects interatomic potentials, constructs and runs simulations, and performs iterative error recovery within a closed-loop workflow. We evaluate the scientific reliability of the agent by benchmarking its outputs against LAVA, a high-throughput toolkit for LAMMPS and the Vienna Ab initio Simulation Package (VASP) calculations. Our framework reduces manual intervention and trial-and-error, thereby improving the rigor, reproducibility, and scalability of atomistic modeling.} \\

\end{tabular}

 \end{@twocolumnfalse} \vspace{0.6cm}

  ]

\renewcommand*\rmdefault{bch}\normalfont\upshape
\rmfamily
\section*{}
\vspace{-1cm}


\footnotetext{\textit{$^{a}$~Los Alamos National Laboratory, Los Alamos, NM, USA}}
\footnotetext{\textit{$^{\dag}$~Email: rsomasundaram@lanl.gov }}



\section{Introduction}
Atomistic simulations, such as molecular dynamics (MD), play a key role in modern materials science by enabling predictive modeling of material properties~\cite{van2020roadmap}. Among available tools,  Large-scale Atomic/Molecular Massively Parallel Simulator (LAMMPS)~\cite{thompson2022lammps} has emerged as a widely used platform due to its flexibility and extensibility. However, effectively using LAMMPS remains a nontrivial task: even routine simulations require careful selection of interatomic potentials, construction of input scripts, parameter tuning, execution management, and post-processing of results~\cite{plimpton1995fast,DANG2023108667}. These steps often demand significant domain expertise and involve iterative trial-and-error, limiting accessibility, reproducibility, and scalability in computational materials research. 

Recent advances in Large Language Models (LLMs)~\cite{naveed2025comprehensive}, scientific foundation models~\cite{bommasani2021opportunities,herde2024poseidon,rautela2025morph,rautela2026pde}, and agent-based systems~\cite{lu2024ai,yamada2025ai,gottweis2026accelerating,ghafarollahi2025sciagents,schmidgall2025agent,grosskopf2025ursa} offer a promising avenue for automating complex scientific workflows. With this aim, the Universal Research and Scientific Agent (URSA) framework~\cite{grosskopf2025ursa} was developed, which is notable for its highly modular and composable system of agents dedicated to autonomously planning, hypothesizing, and executing computational tasks. However, for simulation workflows as complex as atomistic simulations, a system that provides a layer of tightly structured orchestration, implemented on top of URSA's autonomous nature, is highly desirable. In this work, we present such an agentic framework for automating LAMMPS simulations within the URSA ecosystem. Our system is designed to autonomously manage the full lifecycle of an MD simulation, including but not limited to, interatomic potential selection, input script generation, execution, error recovery, and post-processing.

\subsection{Related Work}
The automation of atomistic simulations via LLM-based agents has previously been studied in the literature. Notable examples for materials modeling are AtomAgents~\cite{ghafarollahi2024atomagents} and the Multi-agentic AI framework of Ref.~\citenum{VRIZA2025440}. Both works emphasize collaboration among multiple agents, where each individual agent has expertise in a specific domain. For instance, AtomAgents focuses on a multi-modal GPT-4 based system of agents, where each agent operates within a well-defined role such as proposing a plan to solve the user-specified problem, generating a hypothesis in the realm of material science, critiquing a plan, etc. In the AtomAgents framework, all atomistic simulations are carried out using human generated LAMMPS scripts. Ref.~\citenum{VRIZA2025440} extends such a multi-agent approach, employing highly specialized agents for tasks such as the calculation of elastic constants of a material, the melting temperature, phonon dispersion, etc. This work also presents agents capable of downloading interatomic potentials from the internet, authoring LAMMPS scripts, visualization, etc. 

Another notable example is MDCrow~\cite{campbell2025mdcrow}, developed for automating biomolecular MD workflows. Furthermore, frameworks such as El Agente~\cite{zou2025agente} and DREAMS~\cite{wang2025dreams} provide similar agent-based automation but for DFT simulations. Clearly, these papers demonstrate the increasingly impactful role played by agentic AI systems in complex scientific tasks such as materials modeling. Our work further contributes to this rapidly evolving field of research by refining certain key aspects. 

\begin{figure*}
    \centering
    \includegraphics[trim = 1.25cm 1.0cm 1.8cm 1.25cm, clip, scale=0.5]{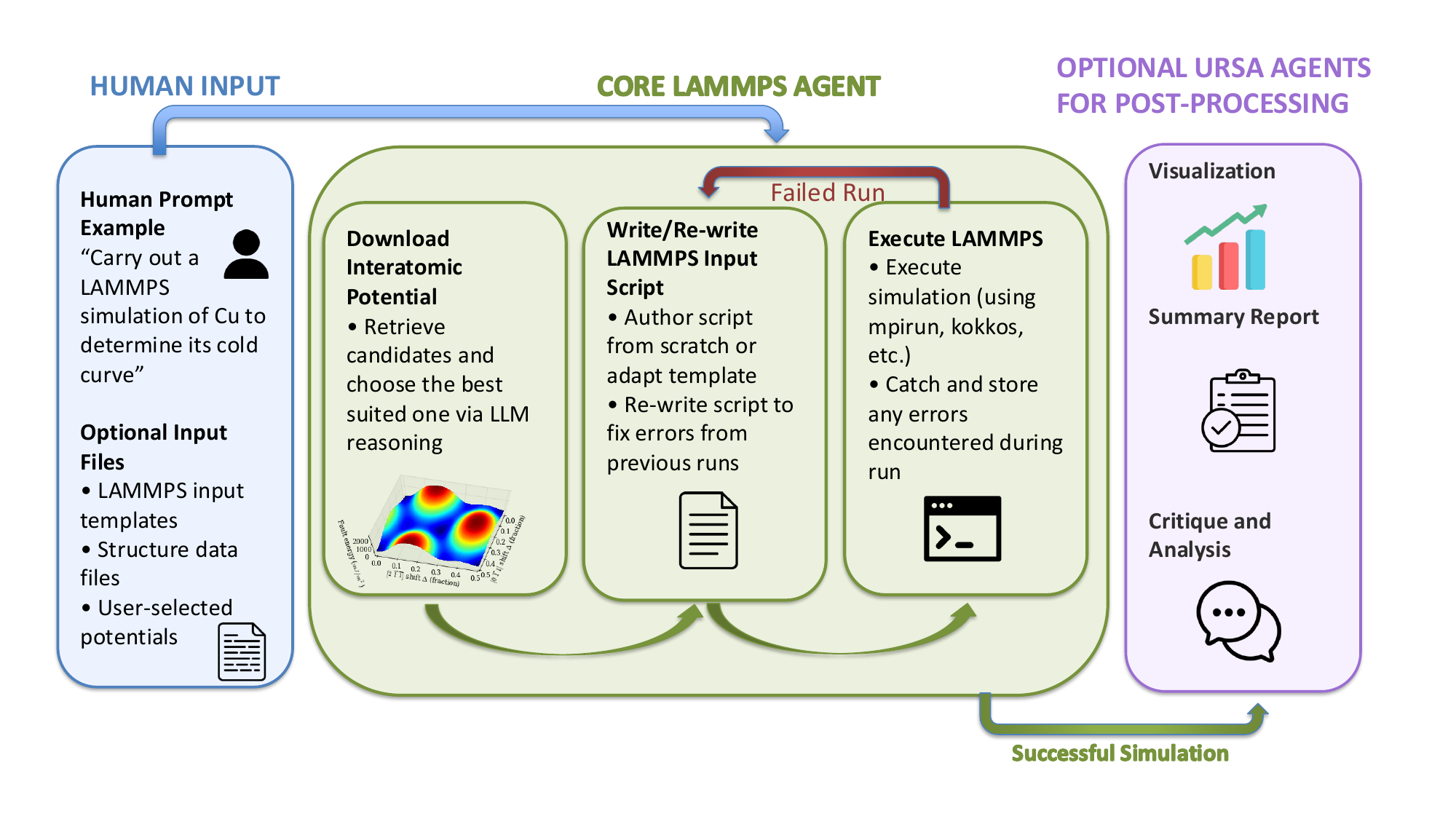}
    \caption{Architecture of our MD agentic workflow}
    \label{fig:architecture}
\end{figure*}

\subsection{Contributions}
\begin{itemize}
    \item In our work, we show that an entire MD workflow can be performed end-to-end, starting from selecting/downloading an interatomic potential to successfully executing LAMMPS, by a single agent. Therefore, we significantly lower the orchestration overhead, for example, by avoiding agent-to-agent handoffs~\cite{wu2024autogen}. This single agent forms the core of our work and is referred to hereafter as our LAMMPS agent (or, equivalently, our MD agent). 

    \item In our LAMMPS agent, the selection and downloading of the interatomic potential is structured and operationalized as a human-reproducible retrieval-and-ranking loop.

    \item Our LAMMPS agent can be operated in two modes: a minimal mode where the agent exhibits relatively more autonomy, suitable for users without MD expertise, as well as an “expert mode” where the agent acts on input from an expert user (LAMMPS templates, specialized prompts, desired potential, etc.). 

    \item Our single agent framework allows for an easy incorporation of a human-in-the-loop. For example, the human can interrupt the LAMMPS agent's workflow and restart it with a different interatomic potential, overriding the agent's choice. 

    \item Finally, while the entire MD workflow is accomplished by a single core LAMMPS agent, this agent can optionally interact with other agents (unrelated specifically to MD) within URSA's ecosystem. These optional agents can be used for visualizing the results of an MD simulation, scanning arXiv for relevant papers, etc. 
\end{itemize}

With these contributions, we believe that our work significantly accelerates the adoption of agentic AI enabled workflows for the design and discovery of novel materials.

\section{Architecture}
\label{sec:arch}

The general schematics of our agent-enabled MD workflow is shown in Fig~\ref{fig:architecture}. The overall scheme can be divided into three parts: the human-specified input that invokes the agent, the core LAMMPS agent, and the optional handoff to other URSA agents for postprocessing. The human input can consist of only a single prompt which specifies the simulation task and launches the workflow. This constitutes a `minimal mode' which involves running the workflow at a higher level of agent-enabled autonomy. Here, for example, the agent chooses the potential for the MD simulation. Alternatively, for users with a higher level of expertise in MD simulations, the agent takes in optional inputs such as potential files for use in the simulation. Notably, the user can also provide a specific template for LAMMPS input scripts that the agent can work off, which might be necessary for specialized MD tasks. 

As with all of URSA's agents, the LAMMPS agent is implemented using LangGraph~\cite{langgraph}, i.e. as a graph-based model which evolves the agent's state. In this work, the agent's state is structured to enable a significant level of human-in-the-loop, as well as persistence across multiple runs. Key variables such as the chosen interatomic potential, any user-provided template for the LAMMPS input script, agent-generated LAMMPS input scripts, any LAMMPS simulation errors, etc., are implemented as the agent's state variables. These variables can therefore change as the agent's state flows through the edges and nodes of the graph. Variables such as the chosen potential can be pre-populated by the user before initiating the agent. 
When the agent is invoked, an entry router is activated which, based on the agent's state, starts the workflow from a specific graph node. For example, if the interatomic potential is present as a state variable, tasks such as autonomously selecting a potential are skipped. Similarly, the user can pre-populate the template variable in the state, using a LAMMPS script from a previous run. In this case, the agent is instructed to work off this template for generating the input deck for the current run.  

\begin{figure*}
    \centering
    \includegraphics[width=0.99\linewidth]{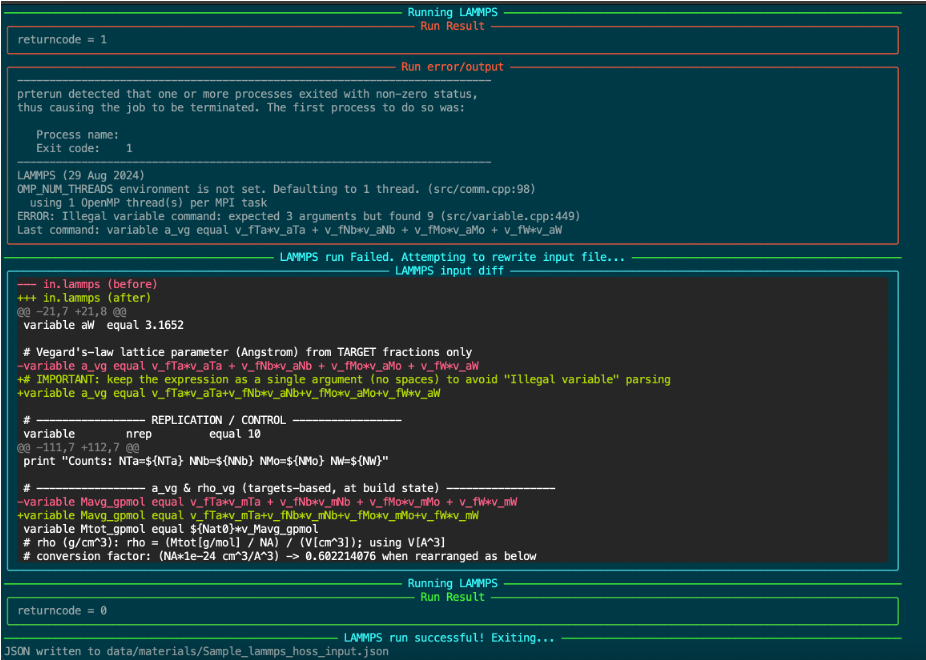}
    \caption{Terminal output during the agent's generation of the simulation's input script. The error originating from an initial attempt is enclosed in a red box. The subsequent edits made by the agent to fix this error are highlighted as `LAMMPS input diff'.}
    \label{fig:input_diff}
\end{figure*}

We describe the algorithm used by the agent to select and download an interatomic potential in detail in Sec.~\ref{sec:pots}. In this section, we focus on the part of the LAMMPS agent that is responsible for authoring and subsequently refining the LAMMPS script. For this, the agent first invokes an LLM that is tasked with authoring the script that is aligned with the user's requested simulation task. With this script, the agent then executes the simulation via MPI, or kokkos~\cite{trott2022kokkos} in the case of GPU-enabled computing resources. If the simulation results in an error, the agent iteratively attempts to fix it by invoking a second LLM and providing it with the full history of all previously generated LAMMPS scripts and their associated errors. This iterative procedure occurs until the simulation is successful or until a user-specified limit on the number of iterations is achieved. As this procedure takes place, the agent's terminal output displays the initial script authored by the agent, and highlights subsequent edits made to this script to fix any errors. An example of this is shown in Fig.~\ref{fig:input_diff}.

The successful completion of the MD simulation is the termination point of the core LAMMPS agent that is responsible for the MD workflow. At this stage, handoffs to other agents that belong to URSA's family of general purpose agents are possible. URSA includes a large system of agents for various research tasks and these are described in detail in Ref.~\citenum{grosskopf2025ursa}. Here, we very briefly summarize the Execution agent and the arXiv agent as they are used in the remainder of this paper. \\
\textbf{Execution Agent:} URSA's Execution agent is designed to handle code and tool-using tasks. For a given general problem, this agent allows an LLM to autonomously select a specific tool and iterate on the tool until the task is achieved. In this paper, we employ this agent's `write code' tool that allows the agent to write python code that can read, summarize, and visualize the simulation log files generated by the LAMMPS agent. \\
\textbf{ArXiv Agent:} This agent uses the arXiv search API (Application Programming Interface) to find papers relevant to a given problem. It then uses LLMs to process the text and images in the papers to summarize their contents in the context of a user-specified task. To accomplish this, the agent can use either the full-texts of the papers, or create and use Retrieval-Augmented Generation (RAG)-based vectorstores for each paper.

\section{Agentic Potential Selection Strategy}
\label{sec:pots}

\begin{figure*}
    \centering
    \includegraphics[trim = 0.25cm 1.75cm 0.25cm 1.75cm, clip,scale=0.5]{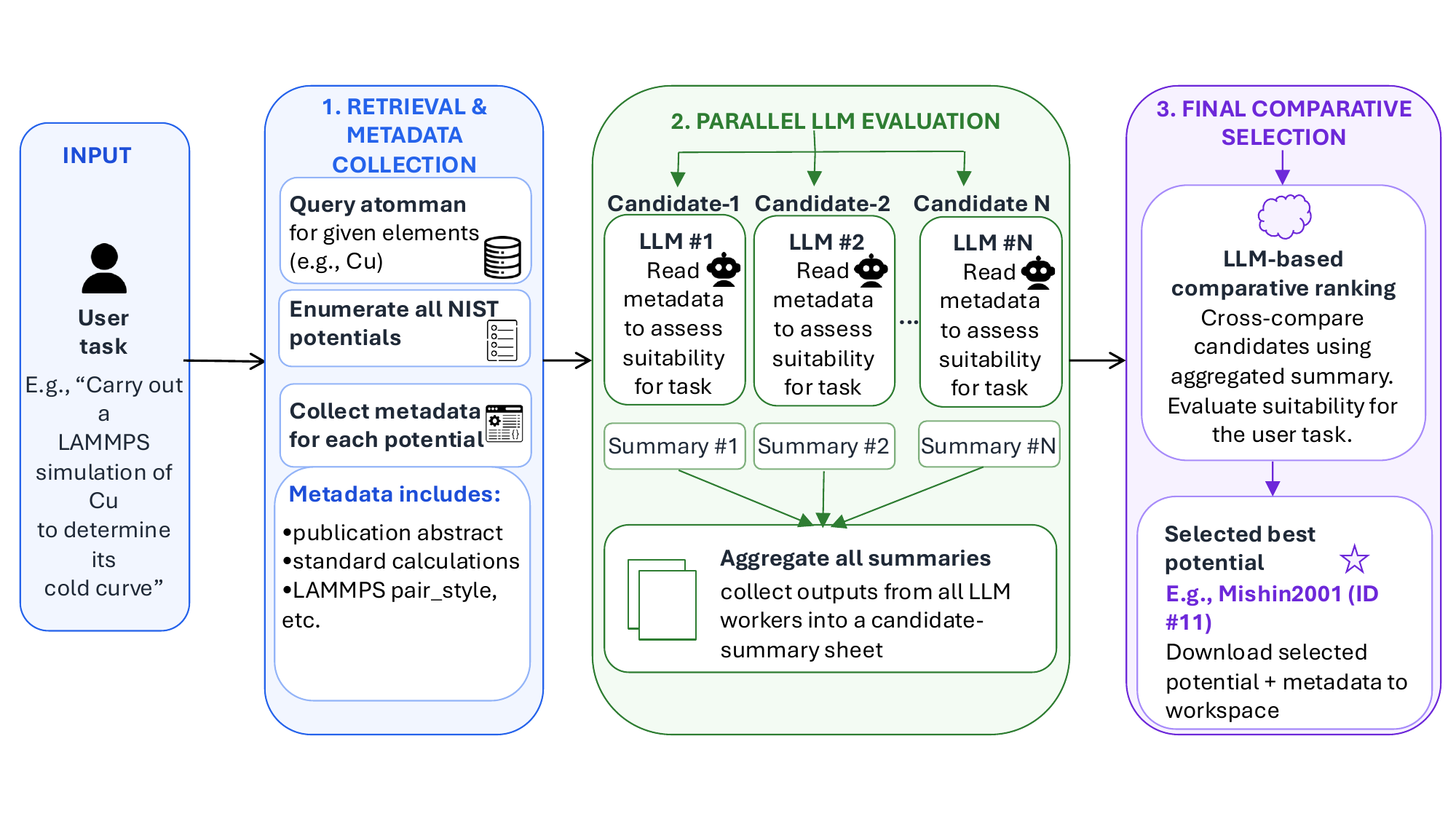}
    \caption{Algorithm for LLM-based potential selection}
    \label{fig:potential_graph}
\end{figure*}

Similar to Ref.~\citenum{VRIZA2025440}, our LAMMPS agent is capable of obtaining interatomic potentials from the internet. However, instead of automated web mining, our framework employs a structured approach that allows the agent to systematically evaluate all relevant potentials in a given database, chosen here to be the National Institute of Standards and Technology (NIST) repository~\cite{nist_ipr}, before choosing one through LLM-based reasoning. The key steps outlining our potential selection strategy are shown in Fig.~\ref{fig:potential_graph}. First, we employ the atomman package~\cite{hale_atomman_2019}
to enumerate all available interatomic potentials for a given set of desired elements. Then, for each potential, the agent obtains all metadata available for the potential from the NIST repository which typically includes information such as the abstract of the corresponding publication, results of standard calculations, LAMMPS implementation information, etc. An ensemble of LLMs are then used, where each LLM is tasked with creating a brief human-readable summary of the usefulness of a single potential for the desired MD simulation, given the aforementioned metadata. The results of all such outputs are then passed to an administrator-like LLM which, based on these outputs, makes the final decision. This potential is then automatically downloaded into the current workspace directory. In this manner, we ensure an automated systematic algorithm that allows a human-in-the-loop to examine the rationale behind the agent's choice over other possible options.

As an illustration, we demonstrate a minimal example where the agent is tasked with choosing an interatomic potential for a LAMMPS simulation of copper. For this experiment, the GPT-5.2 model~\cite{openai_models_2025} is used. The agent is initiated with the prompt:
\begin{tcolorbox}[colback=blue!5!white]
Carry out a LAMMPS simulation of Cu to determine its cold curve.
\end{tcolorbox}

For copper (Cu), there are $101$ potentials in the NIST repository. In our framework, this involves $101$ independent LLM calls, and the output of one such LLM is shown as an example in Appendix~\ref{sec:pot_example}. 
Finally, in this example, the administrator LLM chooses the potential of Ref.~\citenum{PhysRevB.63.224106} with the following rationale

\begin{tcolorbox}[colback=blue!5!white]
\section*{Chosen potential \#11}

\textbf{Rationale for choosing this potential:}

Mishin et al.\ (2001) is a widely validated, Cu-focused EAM potential with a stable fcc 0~K ground state and well-documented equilibrium properties, and it is provided in a directly usable LAMMPS format (\texttt{pair\_style eam/alloy}). That combination makes it a reliable and convenient choice for generating a smooth and physically reasonable Cu energy--volume (cold-curve) dataset via systematic lattice/volume scaling and 0~K minimization in LAMMPS.
\end{tcolorbox}

For comparison, we used the potential agent of Ref.~\citenum{VRIZA2025440}, initiated with the same prompt above, to select and download a potential for this task. We found that this agent selected the 1986 EAM potential of Ref.~\citenum{foiles1986embedded}, while our framework selected the later Mishin et al. potential of Ref.~\citenum{PhysRevB.63.224106}. While both potentials are suitable for copper simulations, the Mishin potential was constructed using a larger fitting database incorporating both experimental and \textit{ab initio} structural energies, and was explicitly benchmarked against deformation paths and nonequilibrium configurations relevant to equation-of-state calculations. The original study further demonstrated improved agreement with \textit{ab initio} and experimental compressed-state energetics relative to more traditional EAM fitting schemes. This illustrates how structured metadata retrieval and comparative LLM-based reasoning can influence the selection of task-specialized interatomic potentials beyond purely web-derived retrieval.

\begin{figure*}
    \centering
    \includegraphics[width=0.99\linewidth]{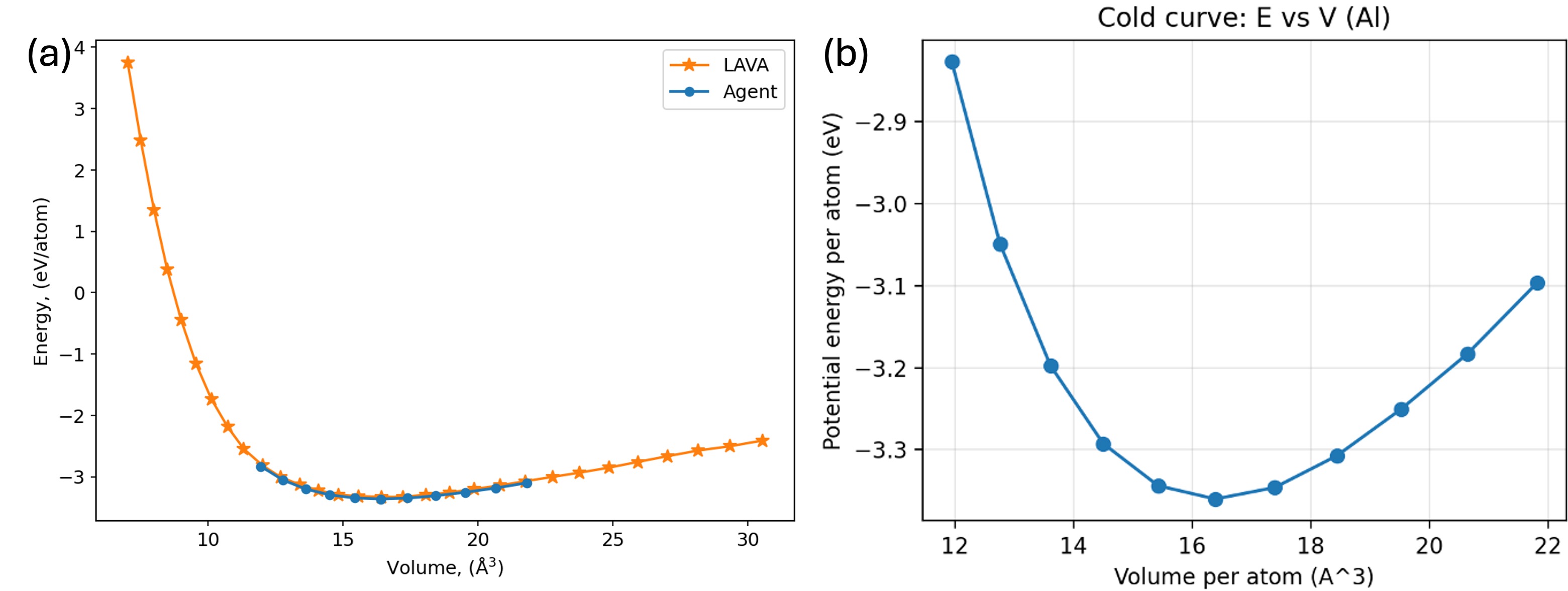}
    \caption{The cold curve of Al computed using LAVA (a) and our agent (b). The agent's results were obtained using the cold-curve prompt given in Appendix~\ref{sec:simple_prompts}. Our agent's output from (b) is plotted in (a) for better visualization and the comparison shows how our agent is able to obtain the correct energies in the relevant range of lattice parameters.}
    \label{fig:cold_curve}
\end{figure*}

\section{Benchmarks}
\label{sec:benchmark}

Our LAMMPS agent, either by autonomously selecting a potential as described in Sec.~\ref{sec:pots}, or by using a user-provided potential, executes LAMMPS for the desired simulation task. As shown below, for standard tasks, this can be accomplished with minimal human input and prompting. For more complex simulations, we show how the agent can adapt to more complex prompts, or alternatively work from user-provided templates for the LAMMPS input deck. 

In this section, we demonstrate these features by benchmarking our agent against LAVA~\cite{DANG2023108667}, a python toolkit for calculations with LAMMPS and the Vienna Ab initio Simulation Package (VASP)~\cite{kresse1993ab}. We choose aluminum (Al) as our benchmark problem and calculate material properties with the potential of Ref.~\citenum{LIU19993227}. For all material properties calculated by the LAMMPS agent, we use URSA's Execution agent to visualize and summarize the results of the simulation, see Sec.~\ref{sec:arch}. For these benchmarks we use the GPT-5~\cite{openai_models_2025} model as the backend LLM. 

\begin{figure}
    \centering
    \includegraphics[width=0.99\linewidth]{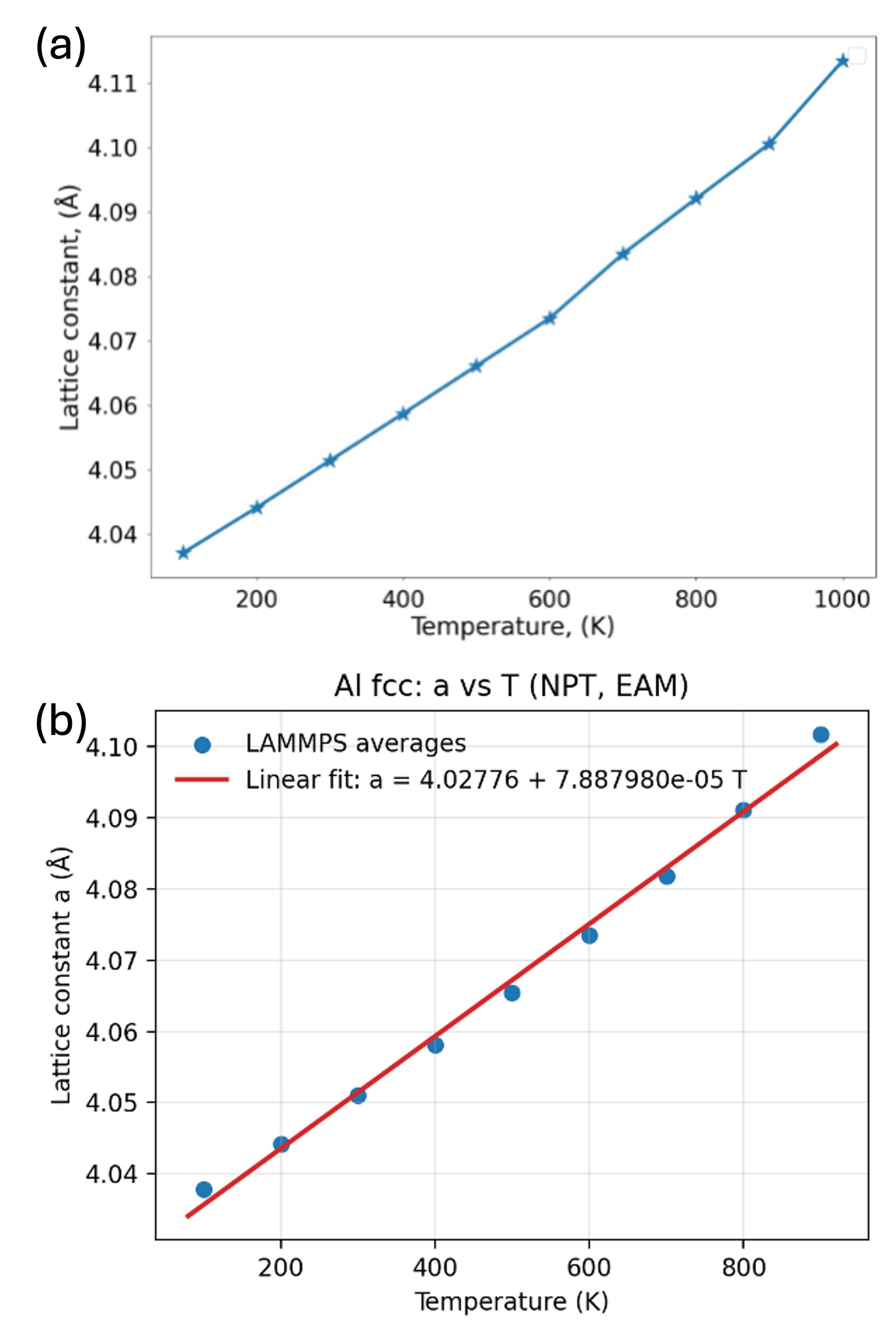}
    \caption{The thermal expansion of Al computed using (a) LAVA and (b) our agent. Results from the agent closely match the calculation done by LAVA. Note that the temperature range our agent picked is smaller than the one set in LAVA.}
    \label{fig:thermal_expansion}
\end{figure}

\subsection{Minimal mode calculations: Cohesive energy, cold curve, and thermal expansion}

First, we present results for material properties of Al that our agent can compute in a minimal mode, i.e. a mode with minimal human input consisting of a very simple prompt specifying the property to be computed. This includes quantities such as the cohesive energy, lattice constant, the cold curve~\cite{birch1947finite}, and the thermal expansion.  
For the cohesive energy, we use the following prompt,
\begin{tcolorbox}[colback=blue!5!white]
Carry out a LAMMPS simulation of Al to determine its cohesive energy.
\end{tcolorbox}
Similar prompts used for the cold curve and the thermal expansion are given in Appendix~\ref{sec:simple_prompts}. Our agentic framework successfully carries out these tasks, and the execution agent provides a detailed summary of the simulations in markdown documents. For the cohesive energy and lattice constant calculation, the key excerpt of such a summary is as follows,

\begin{tcolorbox}[colback=blue!5!white]
\section*{Key Results}
\begin{itemize}
    \item Equilibrium lattice constant $a_0$: 4.03207334~\AA
    \item Cohesive energy: $-3.36000675$~eV/atom
    \item Number of atoms in cell: 4
    \item Potential: 1999--Liu-X-Y--Al-Cu--LAMMPS--ipr1 (al-cu-set.eam.alloy)
    \item Reference: https://doi.org/10.1016/s1359-6454(99)00186-x
\end{itemize}
\end{tcolorbox}
The corresponding main results from LAVA are $a_0 = 4.0321$~\AA, with a cohesive energy of $-3.36$~eV/atom, which is in complete agreement with the agent's calculation. Similarly, our benchmarking results for the cold curve and thermal expansion are shown in Figs.~\ref{fig:cold_curve} and~\ref{fig:thermal_expansion}, respectively. In both cases, the figure for the LAMMPS agent's simulation is produced by the execution agent. In panel (b) of Fig.~\ref{fig:thermal_expansion}, the blue points are results of the simulations performed by the LAMMPS agent whereas the linear fit (red line) was performed by the execution agent when tasked with summarizing and visualizing the results of the LAMMPS agent's simulation. For both the cold curve and the thermal expansion experiment, we find excellent agreement between LAVA and our agentic results. 

\subsection{Expert mode calculations: liquid RDF and melting temperature}
\label{sec:RDF_melting}

\begin{figure*}
    \centering
    \includegraphics[width=0.99\linewidth]{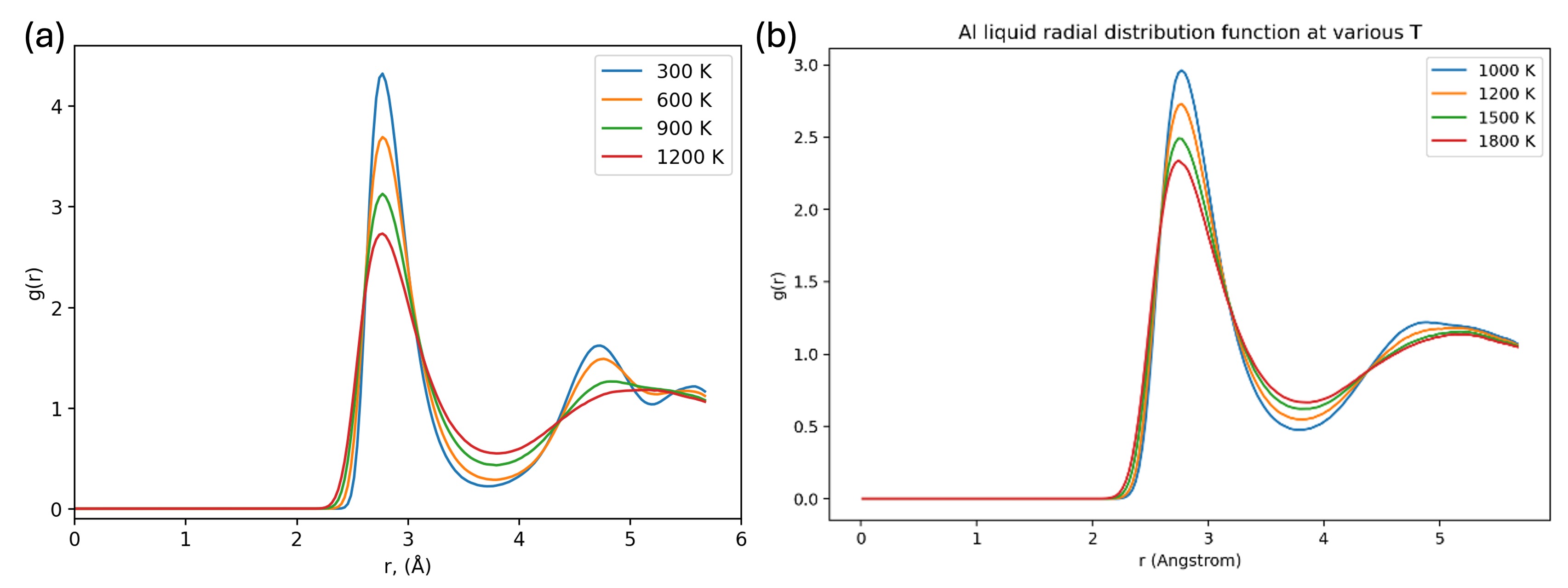}
    \caption{Liquid RDF of Al computed using (a) LAVA  and (b) our agent. The agent's results were obtained using the `minimal prompt' specified in Sec.~\ref{sec:RDF_melting}. Note that the agent, in panel (b), picked a range of temperatures higher than the one used in LAVA and therefore the  interatomic peaks that appear around 4.7$\textit{\AA}$ differ from the ones calculated by LAVA.}
    \label{fig:RDF}
\end{figure*}

We now move towards properties where more complex user instructions are required in order for the agent to achieve the desired result. First, we focus on the liquid radial distribution function (RDF) of Al. 
As with the previous examples, Fig.~\ref{fig:RDF} shows the comparison between LAVA and our agent for the liquid RDF of Al. Interestingly, we see that while the two are consistent, the agent performs calculations only for temperatures above $1000$~K. We attribute this to the fact that the temperature range was not specified explicitly in the minimal prompt, which in this case was,
\begin{tcolorbox}[colback=blue!5!white]
Carry out a LAMMPS simulation of Al to determine its liquid radial distribution function with varying temperature. Use 4.03207334 Angstrom as the lattice constant in the calculation.
\end{tcolorbox}

\begin{figure}
    \centering
    \includegraphics[width=0.99\linewidth]{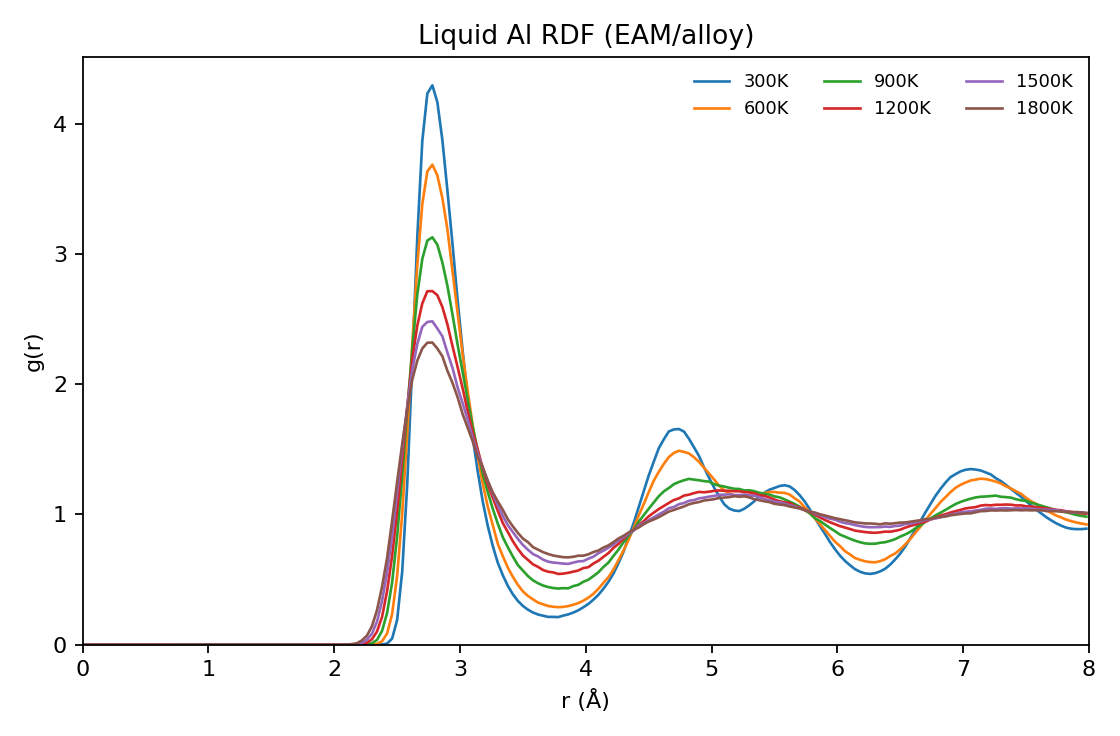}
    \caption{Our LAMMPS agent's results for the liquid RDF of Al with a user-specified temperature range (the detailed prompt specified in Sec.~\ref{sec:RDF_melting}), showing improved agreement with LAVA.}
    \label{fig:RDF_agent_better_prompt}
\end{figure}

\begin{figure}
    \centering
    \includegraphics[width=0.99\linewidth]{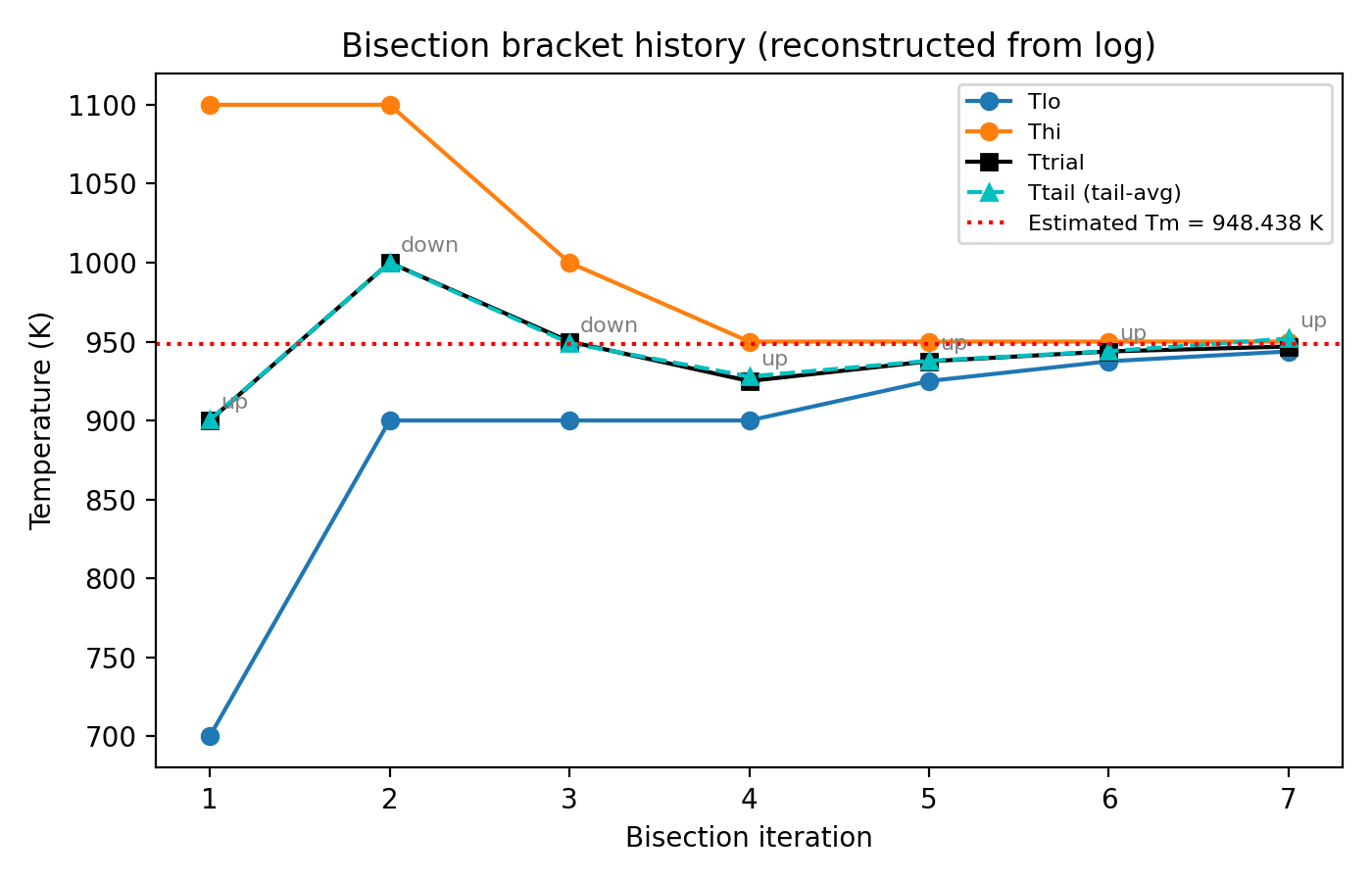}
    \caption{Our agentic results for the bisection-based determination of the melting temperature of Al, showing convergence of temperature brackets across iterations. These results match well with the corresponding LAVA calculations; see the text for details.}
    \label{fig:melting_agent}
\end{figure}

We have found that in such cases where the results of the agent are not consistent with the user's expectations, the agent's performance can be refined through more detailed instructions. For example, the prompt below specifies details such as the temperature grid and produces the results shown in Fig.~\ref{fig:RDF_agent_better_prompt}, where the temperature range is consistent with the LAVA calculations.
\begin{tcolorbox}[colback=blue!5!white]
Simulate liquid aluminum. Start from an FCC Al crystal, melt it at high temperature under NPT conditions at zero pressure, then study the liquid structure over a range of temperatures. For each target temperature between 300 K and 1800 K (in 300 K increments), first equilibrate the liquid under NPT at zero pressure to establish the correct density, then switch to NVT at fixed density and compute the radial distribution function (RDF). Output one RDF file per temperature and ensure the simulation is stable, physically reasonable, and free of common LAMMPS RDF cutoff/ghost atom errors.
\end{tcolorbox}
This demonstrates that our agent is capable of adapting to complex user instructions that are necessary for specific MD simulations. Next, we study the melting temperature. 
Calculations of the melting temperature via MD simulations are non-trivial since different algorithms exist to calculate this quantity, and they typically lead to different results. In LAVA the melting temperature is calculated using the two-phase (or solid-liquid coexistence) method~\cite{morris1994melting}. With LAVA, for the given interatomic potential, we find a melting temperature of $934.4 \pm 20$~K. First, we benchmark our LAMMPS agent against this result first by employing the minimal melting-temperature prompt shown in Appendix~\ref{sec:prompts_melting}.
With this prompt, the agent calculated a melting temperature of $868 \pm 5$~K. We attribute this discrepancy, relative to our LAVA result, to the lack of the LLM's knowledge about the two-phase method. However, as with the RDF calculation, we found that the agent was able to respond effectively to a more descriptive prompt where the details of the two-phase method are elaborated (see Appendix~\ref{sec:prompts_melting} for the detailed coexistence-and-bisection prompt).
In this case, we find that the melting temperature lies within the bracket $[946.875, 950]$~K, which is now consistent with LAVA's prediction. The execution agent's summary of the LAMMPS simulation includes Fig.~\ref{fig:melting_agent} where the progression of the LAMMPS simulations via the bisection algorithm is shown. 

This melting-temperature example highlights a conceptual difference between our agentic framework and earlier multi-agent approaches that employ dedicated agents implementing predefined workflows. For example, in Ref.~\citenum{VRIZA2025440}, melting-temperature calculations are handled through a specialized agent whose system-level instructions prescribe a particular simulation protocol for constructing and evolving a solid–liquid interface. While such workflow-specific agents can effectively automate established procedures, the numerical methodology is largely encoded within the agent implementation itself. Note that this workflow differs from the coexistence-and-bisection methodology implemented in LAVA and reproduced in our benchmarks above. In contrast, our framework exposes the simulation methodology to the user through prompts and templates, allowing the algorithm to be modified dynamically without requiring new agent implementations. As demonstrated above, the same LAMMPS agent can adapt from minimal prompting to detailed user-specified coexistence and bisection strategies for melting-temperature determination. We believe this enables greater methodological flexibility which is generally reflected in the simulation protocols generated by the two frameworks. To illustrate this in a more concrete manner, we compare the LAMMPS input decks generated when both systems are provided with the same detailed prompt. The LAMMPS input deck generated by the framework of Ref.~\citenum{VRIZA2025440}, when initiated with the same detailed melting-temperature prompt used in this work, is given in Appendix~\ref{sec:their_agent}. We see that the input decks from the two agentic systems share similarities but also differ critically when preparing the structure including the solid-liquid interface. For example, the framework of Ref.~\citenum{VRIZA2025440} builds a cell of size [66.4683 $\text{\AA}$, 66.4683 $\text{\AA}$, 132.937 $\text{\AA}$] and adds the \textit{fix setforce 0.0 0.0 0.0} command to freeze the solid section while melting the other half. In comparison, our agent builds a cell size of [48.6 $\text{\AA}$, 48.6 \text{\AA}, 97.2 \text{\AA}] and applies NPT to the whole box at 2000 K with only a thermostat on the hot region. As for completion of the actual melting temperature simulation, the agent from Ref.~\citenum{VRIZA2025440} sets the number of steps to be 4$\times$ greater than our agent for each iteration, which significantly slows the simulation, particularly when multiple corrections are required in case of errors in the input deck. 
These differences demonstrate that simulation protocols can vary substantially across agentic frameworks, even when pursuing the same scientific objective. This underscores the need for rigorous benchmarks that evaluate both final predictions and the workflows used to obtain them.

\subsection{Full expert mode: the generalized stacking fault energy}

\begin{figure}[t]
    \centering
    \includegraphics[width=0.99\linewidth]{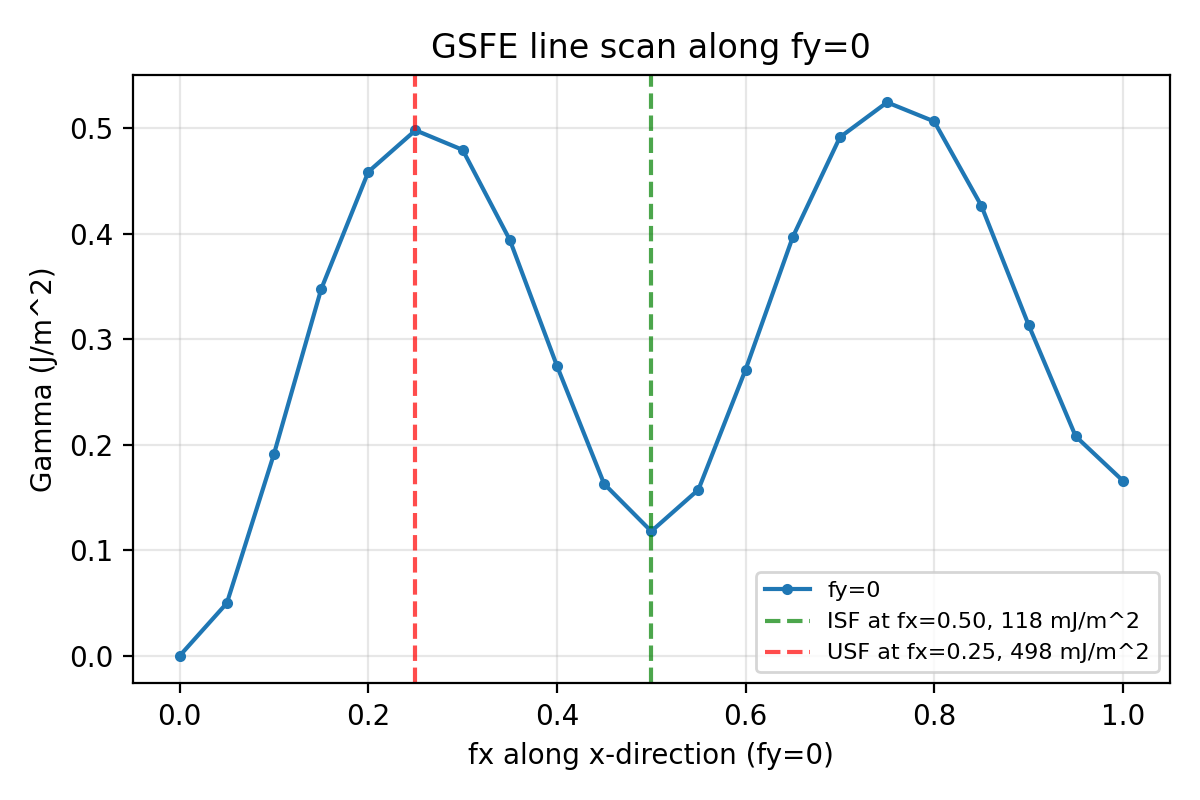}
    \includegraphics[width=0.99\linewidth]{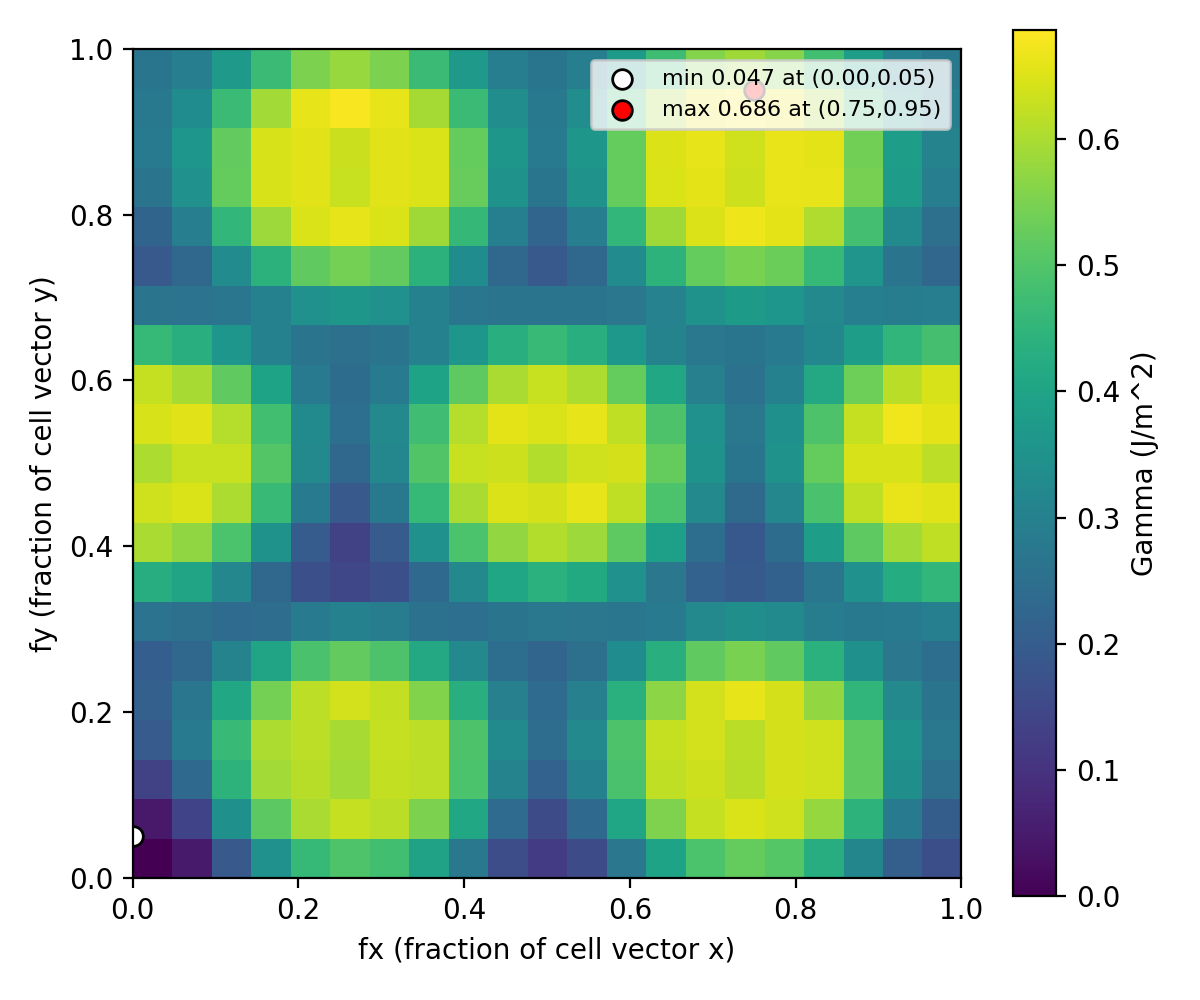}
    \caption{The GSFE surface of Al computed using the LAMMPS agent in `expert mode,' i.e., with a user-provided template. We found that the resulting GSFE profile is fully consistent with the corresponding LAVA calculation, demonstrating that our agent can faithfully reproduce the LAVA workflow when provided with an appropriate simulation template.}
    \label{fig:GSFE_agent}
\end{figure}

We finally present results for the generalized 2D stacking fault energy (GSFE), a property for which the agent is required to be run in full `expert mode'.
For the GSFE, we found that the agent was not able to reproduce LAVA's calculations with a minimal prompt directing the agent to simply calculate this quantity. However, we found that the agent was able to efficiently incorporate a user-provided template for the LAMMPS input script where the calculation of the GSFE was outlined. In general, such templates can be obtained from previous successful runs of the LAMMPS agent. In this case, we generated the template from LAVA. 
Our agent’s results are summarized in Fig.~\ref{fig:GSFE_agent}, which demonstrates the agent’s ability to incorporate the algorithm encoded in the template to compute the GSFE. We found that the resulting GSFE profile is fully consistent with the corresponding LAVA calculation.

We end this section by noting that, although we have presented results here only for a few selected quantities, we have benchmarked our agent against LAVA for all properties that LAVA can calculate. This includes elastic properties (e.g. bulk modulus), 1D and 2D Bain path, Vacancy and interstitial formation energy, etc. For all these quantities, we found that the agent was able to reproduce LAVA's calculation either via minimal prompting or using LAVA generated templates.  

\section{Post-Processing and Validation with Auxiliary URSA Agents}
\label{sec:arxiv}

As discussed in Sec.~\ref{sec:arch}, URSA's ecosystem of agents can be used in a modular manner along with the LAMMPS agent. All figures in Sec.~\ref{sec:benchmark} corresponding to results from our agentic workflow were generated through URSA's execution agent. Here, we further demonstrate the modular nature of URSA's agents by tasking the execution agent to call the arXiv agent as a tool to critique the core LAMMPS agent's calculation of the melting temperature. The prompt we employed for this purpose is given in Appendix~\ref{sec:prompt_arxiv}.
Within the detailed summary and critique of the LAMMPS agent's simulation (output as a markdown file), we find the following excerpts related to the literature review performed by the arXiv agent. 

\begin{tcolorbox}[colback=blue!5!white]
\section*{arXiv-based Validation (Literature Cross-Check)}

An arXiv search for MD/EAM Al melting references returned a useful benchmark via a paper that cites the \textbf{Mendelev et al.\ (2008) Al EAM} potential.

From the arXiv agent summary (based on ArXiv:1512.07891v1 and its cited Al EAM model):

\begin{itemize}
    \item \textbf{Reported EAM Al melting temperature:} $\sim 933~\text{K}$ near low pressure ($\approx 0.2~\text{GPa}$).
\end{itemize}

\subsection{Comparison}

\begin{itemize}
    \item This simulation: $T_m \approx 940.6~\text{K}$
    \item Literature (Mendelev-2008-type Al EAM): $T_m \approx 933~\text{K}$
    \item Difference: $\sim 7$--$8~\text{K}$ ($\approx 0.8\%$)
\end{itemize}

Given typical uncertainties from finite-size effects, interface preparation, and differences between EAM parameterizations, a $\sim 8~\text{K}$ offset is \textbf{not alarming} and is consistent with a reasonable coexistence-style calculation.
\end{tcolorbox}

As can be seen from the output, the arXiv agent was able to find a useful reference in the literature against which the agent's LAMMPS simulation can be verified. This demonstrates how URSA can be used to carry out not just complex numerical simulations, but also other crucial aspects of the workflow involving validation against existing calculations available in the literature.   

\section{Discussion and outlook}

In this work, we have presented URSA's LAMMPS agent--a single agent that is capable of autonomously executing an MD simulation workflow end-to-end. Among its key features, the agent implements a principled LLM-based reasoning strategy to select an interatomic potential for the required MD simulation. The agent can operate with minimal human input (requiring simple prompts indicating the desired quantity to be computed), or with more complex user-defined task specifications and input templates. By incorporating human-in-the-loop, persistence across runs, and a robust iterative strategy to fix LAMMPS input errors, our agent is capable of reproducing the MD results of LAVA across a range of benchmark material properties. Finally, we demonstrated the utility of URSA’s execution and arXiv agents in postprocessing MD simulations through tasks such as summarization, visualization, and literature review.

This work opens up several promising areas of research for further developing the use of agentic AI in materials modeling research. For example, the LAMMPS agent presented here can be interfaced with an optimization agent which can optimize the property of a material with respect to the desired input features. Tackling optimization and inverse problems in this manner using agentic AI methods is a promising research direction for accelerating scientific discovery. Another exciting research direction lies in scale bridging techniques, where simulations across multiple length scales are required to study materials. For instance, the LAMMPS agent presented here could be connected to an agent responsible for handling simulation codes at the mesoscale or continuum scale, such as the HOSS code~\cite{osti_1648095}. Work along these lines is in progress.   

\section*{Conflicts of interest}
There are no conflicts to declare.

\section*{Data availability}
Code availability: all results presented in this work were obtained using the open-source URSA framework that is publicly available \href{https://github.com/lanl/ursa}{here}. This includes the LAMMPS agent discussed in this paper. 

The data associated with this work, such as the configuration files generated by our agent and output log files, are available upon request.

\section*{Acknowledgements}

This work was supported by the Laboratory Directed Research and Development program of Los Alamos National Laboratory under project number 20250638DI. This research
used resources provided by the Los Alamos National Laboratory Institutional Computing Program, which is supported by the U.S. Department of Energy National Nuclear Security Administration under Contract No. 89233218CNA000001.

\bibliography{rsc} 

\bibliographystyle{rsc} 

\appendix

\onecolumn

\section{LLM output for potential selection}
\label{sec:pot_example}

For the experiment in Sec.~\ref{sec:pots}, we show the output of LLM \#50 as an example, instead of the outputs of all $101$ LLMs

\begin{tcolorbox}[colback=blue!5!white]
This Cu potential (Borovikov--Mendelev--King ``fictional Cu-31'', EAM/FS for LAMMPS) is useful for generating a \textbf{cold curve} (0~K energy vs.\ volume or lattice parameter) because:

\begin{itemize}

\item \textbf{It is explicitly packaged for LAMMPS (\texttt{pair\_style eam/fs})} and has already been exercised in NIST's automated property calculations, so it is straightforward to reproduce energy--volume scans reliably.

\item \textbf{It gives a well-defined 0~K fcc reference state for Cu} (A1 fcc relaxed lattice parameter $\approx 3.6386~\mathrm{\AA}$, cohesive/potential energy $\approx -3.4067~\mathrm{eV/atom}$), which is exactly the baseline needed to anchor a cold-curve calculation.

\item The NIST dataset shows it behaves sensibly under \textbf{uniform scaling / energy-vs-distance and energy-vs-$r$ scans}, which are closely related to the workflow for a cold curve (compute minimized energy at a sequence of imposed volumes).

\item Supporting computed properties (elastic constants, surfaces, defects, stacking faults, melting point) indicate the potential is stable enough in typical Cu configurations to allow \textbf{robust minimizations} over a range of compressions/expansions---important when scanning volumes.

\end{itemize}

\textbf{Caveat:} it is a ``fictional'' Cu potential tuned to study SFE/USFE effects, so while it is technically suitable for producing a cold curve, the resulting equation of state may not be the best quantitative match to experiment or DFT. It is therefore best used for methodological testing or for consistency within that potential family.
\end{tcolorbox}

\section{Prompts for the cold curve and thermal expansion}
\label{sec:simple_prompts}

In Sec.~\ref{sec:benchmark}, we employed the following prompts for calculating the cold curve and thermal expansion coefficient of Al.
\begin{tcolorbox}[colback=blue!5!white]
Carry out a LAMMPS simulation of Al to determine its cold curve which is energy vs. (Volume/Density). Use 4.03207334 Angstrom as the lattice constant in the calculation.
\end{tcolorbox}

\begin{tcolorbox}[colback=blue!5!white]
Carry out a LAMMPS simulation of Al to determine its thermal expansion which is lattice constant vs. temperature curve. Use 4.03207334 Angstrom as the lattice constant in the calculation.
\end{tcolorbox}

The value of $4.03207334$ Angstrom for the lattice constant is taken from the agent's calculation of the cohesive energy. 

\section{Prompts and generated input scripts for the melting temperature}
\label{sec:prompts_melting}

\subsection{Melting-temperature prompts}

The following minimal and detailed prompts were used in Sec.~\ref{sec:RDF_melting} for the melting temperature of Al,
\begin{tcolorbox}[colback=blue!5!white]
Carry out a LAMMPS simulation of Al to determine its melting point using 2-phase method.
\end{tcolorbox}

\begin{tcolorbox}[colback=blue!5!white]
Estimate the melting temperature of aluminum using a two-phase coexistence method with bisection. Start from an FCC Al crystal (a0=4.05), split the box into solid and liquid halves, melt only the liquid half at elevated temperature under NPT (zero pressure) and then cool it back to a trial temperature. For each bisection iteration, run a long NPH production at zero pressure and use the tail-averaged temperature drift to decide whether the trial temperature is above or below the melting point (if tail-average T $\geq$ T$_\text{trial}$, move the lower bracket up; otherwise move the upper bracket down). Continue until the bracket width is within a specified tolerance, write a bracket history file each iteration, dump trajectories for the NPH segment, and ensure the script is stable, physically reasonable, and avoids common LAMMPS issues (e.g., thermostat/barostat misuse, bad region/group definitions after clear, and missing log/cleanup between iterations).
\end{tcolorbox}

\subsection{Scripts generated by our framework}

For these prompts, the following LAMMPS input scripts were generated by our agentic framework,

\begin{tcblisting}{
  colback=blue!5!white,
  colframe=blue!50!black,
  title={LAMMPS input script with the minimal prompt},
  fonttitle=\bfseries,
  breakable,
  listing only,
  listing options={
    basicstyle=\ttfamily\normalsize,
    breaklines=true,
    columns=fullflexible,
    keepspaces=true,
    showstringspaces=false
  }
}
log ./log.lammps

units          metal
atom_style     atomic
boundary       p p p

variable       a equal 4.03207334

# Build perfect fcc Al crystal
lattice        fcc ${a}
region         box block 0 12 0 12 0 24 units lattice
create_box     1 box
create_atoms   1 box

mass           1 26.982

pair_style     eam/alloy
pair_coeff     * * ./al-cu-set.eam.alloy Al

neighbor       2.0 bin
neigh_modify   delay 0 every 1 check yes

timestep       0.001

thermo         1000
thermo_style   custom step temp press pe ke etotal density lx ly lz

# Step 1: Equilibrate crystalline solid at 300 K and 0 bar
velocity       all create 300.0 12345 mom yes rot yes dist gaussian
fix            equil all npt temp 300.0 300.0 0.1 iso 0.0 0.0 1.0
run            10000
unfix          equil

# Define two halves along z (lower = solid, upper = to be melted)
region         lower block 0 12 0 12 0 12 units lattice
region         upper block 0 12 0 12 12 24 units lattice
group          solid region lower
group          melt  region upper

# Temperature diagnostics for each half
compute        Tsolid solid temp
compute        Tmelt  melt  temp
thermo_style   custom step temp c_Tsolid c_Tmelt press pe ke etotal lx ly lz

# Step 2: Melt the upper half while holding the lower half fixed
fix            hold   solid setforce 0.0 0.0 0.0
velocity       solid  set 0.0 0.0 0.0 sum yes

# Ramp melt region from 300 K to 2000 K
fix            hot1   melt nvt temp 300.0 2000.0 0.1
run            20000
unfix          hot1

# Maintain at 2000 K to ensure complete melting of upper half
fix            hot2   melt nvt temp 2000.0 2000.0 0.1
run            20000
unfix          hot2

# Step 3: Release constraints and bring the two-phase system to a target guess temperature
unfix          hold

variable       Tguess equal 930.0
fix            eq2    all nvt temp ${Tguess} ${Tguess} 0.1
run            20000
unfix          eq2

# Step 4: Two-phase coexistence in NPH at 0 pressure; observed T ~ Tm
reset_timestep 0
fix            coex   all nph iso 0.0 0.0 1.0
run            200000

# Report final instantaneous temperature; average of thermo output during NPH is the melting temperature estimate
variable       Tfinal equal temp
print          "Estimated melting temperature from two-phase NPH run (instantaneous): ${Tfinal} K"
\end{tcblisting}

\begin{tcblisting}{
  colback=blue!5!white,
  colframe=blue!50!black,
  title={LAMMPS input script with the detailed prompt},
  fonttitle=\bfseries,
  breakable,
  listing only,
  listing options={
    basicstyle=\ttfamily\normalsize,
    breaklines=true,
    columns=fullflexible,
    keepspaces=true,
    showstringspaces=false
  }
}
log ./log.lammps

units metal
dimension 3
boundary p p p
atom_style atomic

# Lattice and box
variable a0 equal 4.05
variable nx equal 12
variable ny equal 12
variable nz equal 24

lattice fcc ${a0}
region box block 0 ${nx} 0 ${ny} 0 ${nz} units lattice
create_box 2 box
create_atoms 1 box

# Masses and potential (after box creation)
mass 1 26.982
mass 2 63.546
pair_style eam/alloy
pair_coeff * * ./al-cu-set.eam.alloy Al Cu

# Neighbor settings
neighbor 2.0 bin
neigh_modify delay 5 every 1 check yes

# Timestep
timestep 0.002

# Computes and thermo
compute Tfull all temp
thermo_style custom step temp pe ke etotal press vol
thermo 200
thermo_modify temp Tfull flush yes

# Define regions/groups for two-phase setup
variable zlo equal zlo
variable zhi equal zhi
variable zmid equal 0.5*(${zlo}+${zhi})
region hot block EDGE EDGE EDGE EDGE ${zmid} EDGE units box
group hot region hot
compute Thot all temp/region hot

# Initialize velocities
velocity all create 300.0 4928459 mom yes rot yes dist gaussian

# Damping parameters
variable tdamp equal 0.2    # ps
variable pdamp equal 2.0    # ps

# --- Melt hot (upper) half under NPT with thermostat using hot-region temperature ---
variable Tmelt_high equal 2000.0
variable melt_steps equal 25000
fix mpt all npt temp ${Tmelt_high} ${Tmelt_high} ${tdamp} iso 0.0 0.0 ${pdamp}
fix_modify mpt temp Thot
thermo 1000
run ${melt_steps}
unfix mpt

# --- Initial melting bracket and cool to initial trial temperature ---
variable Tlo equal 700.0
variable Thi equal 1100.0
variable Ttol equal 5.0
variable maxiter equal 20

variable Ttrial equal 0.5*(${Tlo}+${Thi})
variable cool_steps equal 10000
fix eq0 all npt temp ${Ttrial} ${Ttrial} ${tdamp} iso 0.0 0.0 ${pdamp}
fix_modify eq0 temp Tfull
run ${cool_steps}
unfix eq0

# Prepare bracket history file
print "# iter Tlo[K] Thi[K] Ttrial[K] Ttail[K] decision" file bracket_history.txt

# Bisection loop
variable iter equal 0
label bisect

variable iter equal ${iter}+1
variable width equal ${Thi}-${Tlo}
print "Bisection iter ${iter}: bracket= [${Tlo}, ${Thi}] width= ${width} K"
if "${width} < ${Ttol} || ${iter} > ${maxiter}" then "jump SELF finish"

# Current trial temperature
variable Ttrial equal 0.5*(${Tlo}+${Thi})

# Short NPT equilibration at Ttrial, P=0 (thermostat on all atoms)
variable npt_equil_steps equal 5000
fix eq all npt temp ${Ttrial} ${Ttrial} ${tdamp} iso 0.0 0.0 ${pdamp}
fix_modify eq temp Tfull
run ${npt_equil_steps}
unfix eq

# NPH production at zero pressure
variable nph_steps equal 50000
variable tail_frac equal 0.5
variable tailSteps equal ceil(${nph_steps}*${tail_frac})
variable dump_every equal 200

fix ph all nph iso 0.0 0.0 ${pdamp}

variable T equal temp
fix ft all ave/time 1 ${tailSteps} ${tailSteps} v_T ave running

# Dump trajectory for this NPH segment
dump d all custom ${dump_every} dump_nph_iter${iter}.lammpstrj id type x y z vx vy vz
dump_modify d sort id

thermo 500
run ${nph_steps}

# Tail-averaged temperature over last tail segment
variable Ttail equal f_ft

# Decision and record-keeping
if "${Ttail} >= ${Ttrial}" then "print '${iter} ${Tlo} ${Thi} ${Ttrial} ${Ttail} up' file bracket_history.txt append yes" else "print '${iter} ${Tlo} ${Thi} ${Ttrial} ${Ttail} down' file bracket_history.txt append yes"
if "${Ttail} >= ${Ttrial}" then "variable Tlo equal ${Ttrial}" else "variable Thi equal ${Ttrial}"

# Cleanup for next iteration
unfix ph
unfix ft
undump d

jump SELF bisect

label finish
variable Tmelt equal 0.5*(${Tlo}+${Thi})
print "# Final bracket [${Tlo}, ${Thi}] K ; estimated Tm = ${Tmelt} K" file bracket_history.txt append yes
write_data two_phase_final.data
\end{tcblisting}

\subsection{Script generated by Ref. 17}
\label{sec:their_agent}

With the same detailed melting-temperature prompt, the framework of Ref.~\citenum{VRIZA2025440} generated the LAMMPS input deck below.

\begin{tcblisting}
{
  colback=blue!5!white,
  colframe=blue!50!black,
  title={LAMMPS input script with the detailed prompt generated by the agentic framework of Ref [17]},
  fonttitle=\bfseries,
  breakable,
  listing only,
  listing options={
    basicstyle=\ttfamily\normalsize,
    breaklines=true,
    columns=fullflexible,
    keepspaces=true,
    showstringspaces=false
  }
}

# ------------------------------------------------------------
# 01_relax_interface.in  (region style fixed: use block with EDGE keyword)
# Purpose: Create solid-liquid interface in FCC Al
# ------------------------------------------------------------

units           metal
atom_style      atomic
boundary        p p p

read_data       al_structure.lmp

# ---------------- Potential ----------------
pair_style      eam
pair_coeff      * * Al_jnp.eam

# ---------------- Simulation settings ----------------
timestep        0.001            # 1 fs
neighbor        2.0 bin
neigh_modify    delay 0 every 1 check yes

# ---------------- Energy minimisation ----------------
min_style       cg
minimize        1e-6 1e-8 10000 100000

# ---------------- Equilibrate perfect crystal ----------------
velocity        all create 300.0 12345 mom yes rot yes dist gaussian
fix             npt_eq all npt temp 300.0 300.0 0.1 iso 0.0 0.0 5.0
thermo          1000
thermo_style    custom step temp press pe etotal vol
run             30000            # 30 ps
unfix           npt_eq

# ---------------- Define half-box regions using global box bounds ----------------
variable        ZLO equal zlo
variable        ZHI equal zhi
variable        ZMID equal 0.5*(${ZLO}+${ZHI})

# use block style with EDGE keyword (robust across LAMMPS versions)
region          top  block EDGE EDGE EDGE EDGE ${ZMID} EDGE units box
region          bot  block EDGE EDGE EDGE EDGE EDGE ${ZMID} units box

group           top    region top
group           bottom region bot

# ---------------- Freeze crystalline top ----------------
fix             freeze_top top setforce 0.0 0.0 0.0

# ---------------- Melt bottom half ----------------
velocity        bottom create 3000.0 54321 mom yes rot yes dist gaussian
fix             melt bottom nvt temp 3000.0 3000.0 0.1
run             50000            # 50 ps at 3000 K
unfix           melt

# ---------------- Cool liquid region ----------------
velocity        bottom scale 1500.0
fix             cool bottom nvt temp 1500.0 1500.0 0.1
run             20000            # 20 ps
unfix           cool

# ---------------- Unfreeze and relax whole system ----------------
unfix           freeze_top

velocity        all scale 1200.0
fix             npt_relax all npt temp 1200.0 1200.0 0.1 iso 0.0 0.0 5.0
run             10000            # 10 ps
unfix           npt_relax

# ---------------- Output interface structure ----------------
write_data      interface_structure.data

# Dump a snapshot for verification
reset_timestep  0
dump            d1 all custom 1000 dump_interface.atom id type x y z
run             1000
undump          d1

dump            dump_final all custom 1 dump.output id type x y z
run             0

print "Interface file written: interface_structure.data"

# ------------------------------------------------------------
# 02_bisect_melting.in  (fix IF command with separate statements)
# ------------------------------------------------------------

units           metal
atom_style      atomic
boundary        p p p

# --- Parameters (override with -var ...) ---
variable  Tlow      equal 850.0
variable  Thigh     equal 1000.0
variable  tol       equal 2.0
variable  maxiter   equal 12
variable  npt_pre   equal 20000
variable  nph_prod  equal 200000
variable  tailfrac  equal 0.5
variable  dt        equal 0.001
variable  Tdamp     equal 0.1
variable  Pdamp     equal 5.0
variable  thermoN   equal 1000
variable  dumpN     equal 1000
variable  seedT     equal 93847

variable  datafile  string "interface_structure.data"
variable  potfile   string "Al_jnp.eam"
variable  histfile  string "bracket_history.txt"

shell  if [ ! -f ${histfile} ]; then echo "#iter Tlow Thigh Ttrial TailT Decision" > ${histfile}; fi

variable i loop ${maxiter}
label loop_iter

variable Ttrial equal 0.5*(${Tlow}+${Thigh})

# ---- fresh sim ----
clear
units   metal
atom_style atomic
boundary p p p
read_data ${datafile}

pair_style eam
pair_coeff * * ${potfile}

timestep ${dt}
neighbor 2.0 bin
neigh_modify delay 0 every 1 check yes

thermo ${thermoN}
thermo_style custom step temp press pe etotal vol density

variable iterstr string ${i}
log log.iter_${iterstr}.lammps

# pre-equil NPT
velocity all create ${Ttrial} ${seedT} mom yes rot yes dist gaussian
fix pre all npt temp ${Ttrial} ${Ttrial} ${Tdamp} iso 0.0 0.0 ${Pdamp}
run ${npt_pre}
unfix pre

# NPH production
fix prod all nph iso 0.0 0.0 ${Pdamp}

dump dtraj all custom ${dumpN} dump.iter_${iterstr}.atom id type x y z vx vy vz

variable tailsteps equal ceil(${tailfrac}*${nph_prod})
variable warm equal ${nph_prod}-${tailsteps}
if "${warm} < 0" then "variable warm equal 0"
run ${warm}

variable avefreq equal ${thermoN}
fix ft all ave/time ${avefreq} 1 ${avefreq} c_thermo_temp ave running start 0 file tail.iter_${iterstr}.dat overwrite
run ${tailsteps}

variable TailT equal f_ft

# ----- Bracket update (split into two if commands) -----
if "${TailT} >= ${Ttrial}" then "variable decision string above_or_equal" else "variable decision string below"
if "${TailT} >= ${Ttrial}" then "variable Tlow equal ${Ttrial}" else "variable Thigh equal ${Ttrial}"

print "${i} ${Tlow} ${Thigh} ${Ttrial} ${TailT} ${decision}" append ${histfile}

undump dtraj
unfix prod
unfix ft

variable width equal ${Thigh}-${Tlow}
print "Iteration ${i} finished, bracket width = ${width} K"
if "${width} <= ${tol}" then "print Converged; jump SELF end_bisect"

next i
jump SELF loop_iter

label end_bisect
print "Final bracket: [${Tlow}, ${Thigh}] K  =>  Tmelt ~ $(0.5*(${Tlow}+${Thigh})) K"
print "FINAL ${Tlow} ${Thigh} $(0.5*(${Tlow}+${Thigh}))" append ${histfile}

dump dump_final all custom 1 dump.output id type x y z
run 0
\end{tcblisting}

\section{Prompt for the Execution/arXiv agents}
\label{sec:prompt_arxiv}

We use the following prompt in Sec.~\ref{sec:arxiv} to initiate an arXiv-based review of the MD simulation of the melting temperature.

\begin{tcolorbox}[colback=blue!5!white]
You are part of a larger scientific workflow whose purpose is to accomplish this task: Carry out a LAMMPS simulation of Al to determine its melting temperature. \\

A LAMMPS simulation has been done and all output files are located in the current workspace directory. The simulation logs are recorded in the file 'log.lammps'. \\

Summarize the outcome of this simulation in a markdown document. Include plots, if relevant. Comment on the quality/accuracy of the simulation. Use the arXiv agent to verify if the simulation results are valid.
\end{tcolorbox}

\end{document}